# Single-motor robotic gripper with three functional modes for grasping in confined spaces

Toshihiro Nishimura[1], *Member, IEEE*, and Tetsuyou Watanabe[1], *Member, IEEE*

*Abstract*—This study proposes a novel robotic gripper driven by a single motor. The main task is to pick up objects in confined spaces. For this purpose, the developed gripper has three operating modes: grasping, finger-bending, and pull-in modes. Using these three modes, the developed gripper can rotate and translate a grasped object, i.e., can perform in-hand manipulation. This in-hand manipulation is effective for grasping in extremely confined spaces, such as the inside of a box in a shelf, to avoid interference between the grasped object and obstacles. To achieve the three modes using a single motor, the developed gripper is equipped with two novel self-motion switching mechanisms. These mechanisms switch their motions automatically when the motion being generated is prevented. An analysis of the mechanism and control methodology used to achieve the desired behavior are presented. Furthermore, the validity of the analysis and methodology are experimentally demonstrated. The gripper performance is also evaluated through the grasping tests.

## I. INTRODUCTION

THIS study presents a novel robotic gripper that can achieve three functional modes using a single motor. The developed gripper is shown in Fig. 1(a). The three motion modes of the gripper are as follows:
1) Grasping mode: closing and opening the fingers to grasp and release an object.
2) Finger-bending mode: bending the finger to rotate the fingertip to rotate a grasped object.
3) Pull-in mode: translating a grasped object.

The main target task of this gripper is to pick up an object from extremely confined spaces. Fig. 1(b) shows a representative target situation in which a target object is placed inside a box in a shelf. To pick up objects from confined spaces, the target object must be handled while changing its position and posture to avoid interference from the environment and other objects. The position and posture of the object can be changed by moving the robotic manipulator, but the manipulator motion needs a large workspace. In-hand manipulation, which changes the position and posture of the object while maintaining the manipulator posture, is preferred because it requires a small workspace. This study challenges to develop a gripper that can achieve in-hand manipulation to translate and rotate a grasped object, in addition to the grasping function, through a single motor. Minimizing the number of motors contributes to a compact gripper design that is effective for operation in confined spaces. As shown in Fig. 1(a), the developed gripper realizes the three modes using a single motor. The modes are activated in sequence through unidirectional motor rotation. Automatic motion switching is realized using two novel self-motion switching (SMS) mechanisms. The SMS mechanisms automatically switch motion modes when the motion being generated is prevented. The fingers are initially extended. The forward motor rotation first activates the finger-bending mode, which causes flexion of the finger. If the finger bends 90° where the bending motion is prevented by the stopper installed in the finger, the gripper switches to the grasping mode to allow the finger to close. The contact force exerted by the grasped object on the fingertips increases as the fingers close during the grasping mode. If the contact force becomes sufficiently large to prevent the closing motion, the mode is switched to the pull-in mode. In this manner, the three modes can be achieved through unidirectional motor rotation. When rotating the motor in the reverse direction, the gripper activates each mode in the same order as in the forward motor rotation (Fig. 1(a)). Fig. 1(c) shows the operation of the developed gripper for picking up an object from the confined space shown in Fig. 1(b). The thin finger structure allowed the fingers to be inserted into the narrow space between the box and the top board of the shelf. The finger-bending, grasping, and pull-in modes were sequentially activated through unidirectional motor rotation. The flexion of the finger in the finger-bending mode enabled the fingertip to reach the target object inside the box. Lastly, the reversed motor rotation extended the fingers in the finger-bending mode, moving the fingers to a suitable position for removing the grasped object outwards. The pull-in function reduced the rotational radius for the object rotation, reducing the risk of collisions between the object and the obstacles.

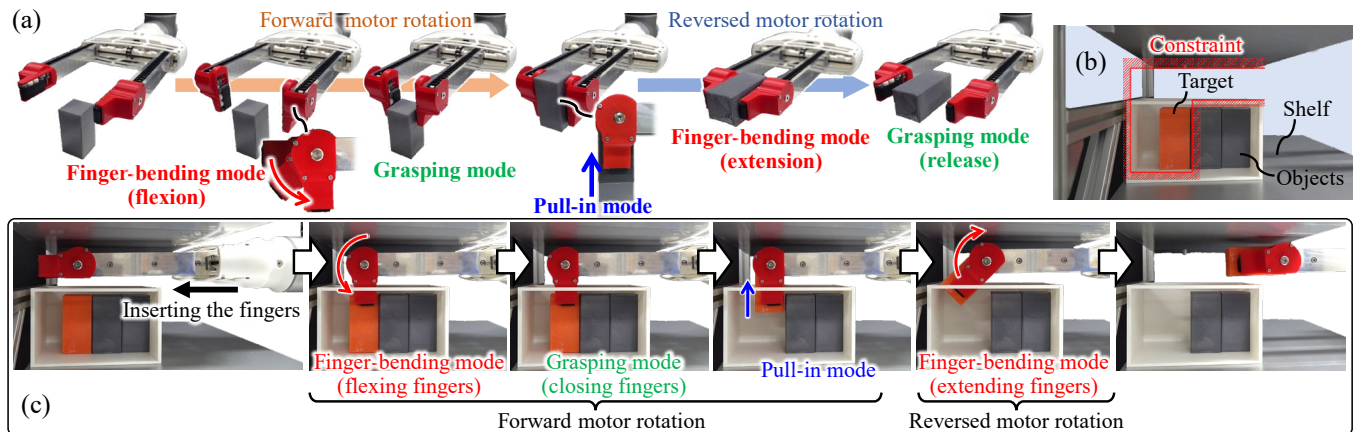

**Fig. 1.** Developed gripper. (a) Gripper operation. (b) Representative target task. (c) Grasping operation in the confined spaces

*A. Related works*

Several studies have focused on grasping objects in confined spaces [1][2]. To accomplish this task, not only control strategies, including object detection and path planning but also end-effectors are required. As described in [3], suction and antipodal grippers are popular for grasping in confined spaces. Suction grippers are effective for insertion into confined spaces because of their slender structure [4][5]; however, it is difficult to stably grasp objects with rough surfaces using suction grippers. Thus, an antipodal gripper is preferable for grasping various objects in confined spaces. A thin structure is also necessary for the grippers to be inserted into confined spaces. Several studies have developed robotic grippers with thin structures [4]–[11]. In the case of extremely confined spaces where the operation range of a manipulator with a gripper is limited, the changing position and posture of grasped objects should be performed in-hand to avoid collisions between the grasped objects and obstacles. It is preferable to be able to both translate and rotate grasped objects. Several studies have attempted to design low-DOF robotic grippers that can change the position or posture of grasped objects through in-hand manipulation [12]-[23]. [12]–[16] realized either translation or rotation of the grasped object. In [17]–[23], both movements were realized by using at least two actuators or environmental supports, such as contact with a supporting surface or gravity. An increase in the number of actuators results in a larger gripper design. Manipulation methodologies that use the environment can be adopted only in limited situations. A 1-DOF robotic gripper that can achieve both translation and rotation of grasped objects without using the environment is preferable. However, no attempts have been made to develop such a gripper.

## II. ROBOTIC GRIPPER DESIGN

*A. Design requirements*

The functional requirements of the developed robotic gripper are as follows: 1) grasping motion and in-hand manipulation, including the rotation and translation of a grasped object, should be achieved using a single motor; and 2) the width of the finger should be equal to or less than 40 mm.

*B. Overview structure*

Fig. 2 shows the three-dimensional computer-aided design (3D-CAD) model of the developed gripper.

1) SMS mechanism A

SMS mechanism A enables switching between the grasping and other modes. To generate the two different motions by a single motor, a screw feed mechanism composed of a screw shaft and nut was adopted. The screw feed mechanism applies the thrust force and axial torque to the nut. Typically, the axial rotation of the nut is restricted and only the translation of the nut is activated. In contrast, both the translation and rotation of the nut can be activaited in SMS mechanism A where the drive shaft (green in Figs. 2 and 3) and sprocket 1 (orange) works as the screw shaft and nut of the feed screw mechanism, respectively. The trasnslation of sprocket 1 generates the grasping motion while the rotation of sprocket 1 activates the other modes, as shown in Fig. 3. As shown in Fig. 2, the drive shaft has an outer thread and is driven by a motor (black) via roller chain 1 (black). Sprocket 1 has an inner thread that engages with the thread of the drive shaft. Sprockets 1 and 2 (yellow) in SMS mechanism B are connected via roller chain 2 (black). The roller chains are used to transmit the motor torque to sprocket 2 through the drive shaft and sprocket 1 to minimize tooth jumping. These sprockets are supported by a finger body (pink) for free axial rotation. The two fingers behave symmetrically because the drive shaft has right-hand and left-hand threads for each finger. The motion of sprocket 1 is determined by the magnitude of the torque applied to sprocket 1 from sprocket 2 through roller chain 2. If the torque is small, sprocket 1 is translated in its axial direction by the screw feed mechanism, activating the grasping mode. If the torque is large, the axial rotation of sprocket 1 occurs, and the rotation of the drive shaft is transmitted to sprocket 2, activating the finger-bending mode or pull-in mode. The generated mode is determined by SMS mechanism B.

2) SMS mechanism B

SMS mechanism B enables switching between the finger-bending and pull-in modes. A belt-conveyor mechanism was adopted to achieve the pull-in mode while the rotation of the roller chain mechanism was used to bend/rotate the fingertip. Fig. 4 shows the detailed structure of SMS mechanism B. The mechanism consists of sprocket 2 with a bevel gear part, and a fingertip unit (Fig. 4(a)). The fingertip unit includes gears 1

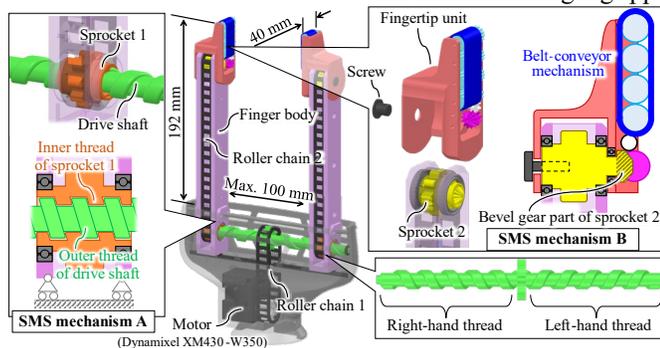

**Fig. 2.** 3D-CAD model of the developed gripper

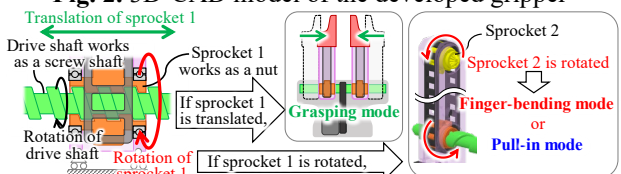

**Fig. 3.** Behavior of SMS mechanism A

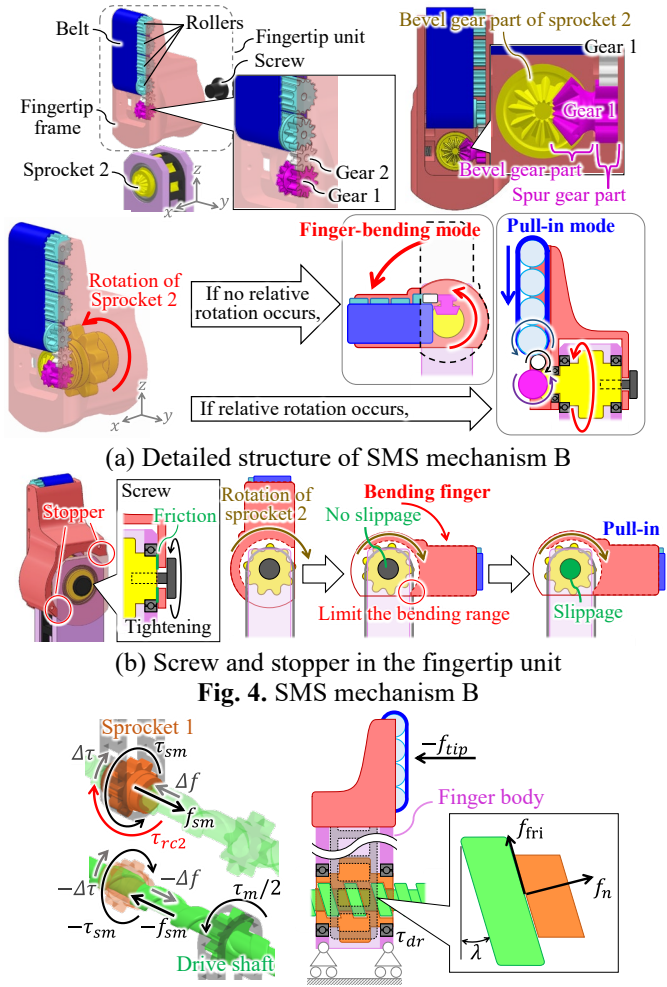

**Fig. 5.** Mechanical relationship in SMS mechanism A (purple) and 2 (white), a belt (blue), rollers (light blue), a screw (black), and a fingertip frame (red). The fingertip unit is attached to sprocket 2 such that they can rotate relative to each other. Gear 1 has a bevel gear part that meshes with the bevel gear part of sprocket 2. Gear 1 has also a spur gear part that meshes with gear 2. The belt and rollers form a belt-conveyor mechanism for the pull-in function. The key motion in SMS mechanism B is the relative rotation between sprocket 2 and the fingertip unit for the motion switching. When sprocket 1 rotates in SMS mechanism A, sprocket 2 also rotates owing to the torque applied by sprocket 1. If sprocket 2 rotates together with the fingertip unit, the finger is bent, and the belt-conveyor mechanism does not work, activating the finger-bending mode. If sprocket 2 rotates while the rotation of the fingertip unit is prevented due to disturbance, relative rotation between them occurs. The relative rotation moves the belt-conveyor mechanism, activating the pull-in mode. To realize self-motion switching through unidirectional motor rotation, a screw and a stopper were installed in the fingertip unit (Fig. 4(b)). The screw is tightened to sprocket 2 to preload the axial force on the contact surface between the fingertip frame and sprocket 2. The preload generates frictional torque on the contact surface to prevent the relative rotation between sprocket 2 and the fingertip unit (frame). Thus, sprocket 2 rotates with the fingertip unit when the rotation of the fingertip unit is not prevented. The stopper in the fingertip frame was designed to limit the range of the finger-bending angle through contact between the stopper and finger body. With the stopper, the finger-bending motion is within the range from the posture where the finger is extended to the posture where the fingertip is perpendicular to the finger body. This mode is maintained until the stopper in the finger unit prevents the rotation of the fingertip unit due to contact between the stopper and finger body. Further motor rotation causes relative rotation between sprocket 2 and the fingertip unit, activating the pull-in mode.

### III. STATICS AND VALIDATION

*A. Statics*

This section analyzes SMS mechanisms A and B to achieve the desired gripper behavior. As shown in Fig. 1(a), the desired behavior is that the unidirectional motor rotation activates the finger-bending, grasping, and pull-in modes in that order. Sections III.A.1 and III.A.2 describe the conditions for mode switching in each SMS mechanism, and Section III.A.3 describes the conditions for achieving the desired order of mode switching. The analysis focuses on one finger because the two fingers in the gripper have a symmetrical structure.

*1) SMS mechanism A*

This section describes the conditions for generating the rotation and translation of sprocket 1 for switching between the finger-bending mode and the other two modes. In this study, the pitch radius of the sprocket installed in the drive shaft, $r_{dr}$, is set to be identical to that of the sprocket attached to the motor. With this setting, the torque applied to the drive shaft from the motor through roller chain 1 is equal to the motor torque $\tau_m$. The mechanical relationship between sprocket 1 and the drive shaft is shown in Fig. 5. From the theory of the lead screw mechanism [24], the thrust force $f_{sm}$ and torque $\tau_{sm}$ applied to sprocket 1 by the screw mechanism are given by

$$f_{sm} = f_n \cos\lambda - f_{\text{fri}} \sin\lambda \quad (1)$$
$$\tau_{sm} = r_{dr}(f_{\text{fri}} \cos\lambda + f_n \sin\lambda) \quad (2)$$

where $f_n$ and $f_{\text{fri}}$ are the normal and friction forces applied to the contact area between the inner thread of sprocket 1 and the outer thread of the drive shaft, respectively; $\lambda$ is the lead angle of the thread. Letting $f_{tip}$ be the external force applied to the fingertip and $\tau_{rc2}$ be the torque applied to sprocket 1 from roller chain 2, the force and moment balances in the axial direction of sprocket 1 are expressed by

$$f_{sm} + f_{tip} + \Delta f = 0 \quad (3)$$
$$\tau_{sm} + \tau_{rc2} + \Delta\tau = 0 \quad (4)$$

where $\Delta f$ and $\Delta\tau$ are the small frictional force and torque exerted on the contact surface between sprocket 1 and the drive shaft except the thread surfaces. $\Delta f$ and $\Delta\tau$ are assumed to be small. $f_{tip}$ is transmitted to sprocket 1 through the fingertip unit and finger body. Assuming that $\tau_m$ is uniformly distributed to the two sprockets 1 owing to the symmetric finger arrangement, the moment balance in the axial direction of the drive shaft is:

$$-\tau_m - 2\tau_{sm} - 2\Delta\tau = 0 \quad (5)$$

From (4) and (5),

$$\tau_m/2 = \tau_{rc2} \quad (6)$$

The quasi-static condition for sprocket 1 to rotate together with the drive shaft without slippage between them is given by

$$|\tau_{sm}|_{f_{\text{fri}}=\mu_{st}f_n} > |\tau_{rc2} + \Delta\tau| \quad (7)$$

where $\mu_{st}$ is the static frictional coefficient. The quasi-static condition in (7) indicates that $\tau_{sm}$ provides the acceleration for the rotation in addition to the torque balancing $|\tau_{rc2} + \Delta\tau|$. Using (3) and (4), the condition is rewritten as

$$|\alpha_{st}r_{dr}(f_{tip} + \Delta f)| > |\tau_{rc2} + \Delta\tau| \quad (8)$$
$$\alpha_{st} = (\mu_{st}\cos\lambda + \sin\lambda)/(\cos\lambda - \mu_{st}\sin\lambda) \quad (9)$$

In contrast, the quasi-static condition required to generate the translation of sprocket 1 is given by

$$|f_{SM}|_{f_{\text{fri}}=\mu_{st}f_n} > |f_{tip} + \Delta f| \quad (10)$$

From (3) and (4), it is rewritten as

$$|(\tau_{rc2} + \Delta\tau)/r_{dr}\alpha_{st}| > |f_{tip} + \Delta f| \quad (11)$$

*2) SMS mechanism B*

This section describes the conditions for generating relative rotation between sprocket 2 and the fingertip unit. The mechanical relationship is shown in Fig. 6. In this paper, the pitch radii of sprockets 1 and 2 are set to be the same; thus, the torque of $-\tau_{rc2}$ is applied to sprocket 2 from sprocket 1 through roller chain 2. The moment balances of sprocket 2 and the fingertip unit are respectively expressed by

$$-\tau_{rc2} + \tau_{pre} + \tau_{g1} = 0 \quad (12)$$
$$-\tau_{pre} - \tau_{g1} + \tau_{stopper} = 0 \quad (13)$$

where $\tau_{pre}$ is the frictional torque between sprocket 2 and the fingertip unit due to the preload by the tightening screw; $\tau_{g1}$ is the torque applied between gear 1 and the bevel gear part of sprocket 2 to drive the belt-conveyor mechanism; and $\tau_{stopper}$ is the torque applied to the fingertip unit due to the contact between the stopper in the fingertip unit and the finger body. Therefore, (12) and (13) are rewritten as

$$\tau_{rc2} = \tau_{stopper} = \tau_{pre} + \tau_{g1} \quad (14)$$

$\tau_{pre}$ increases as $\tau_{rc2}$ and $\tau_{stopper}$ increase. If $\tau_{pre}$ is less than the maximum static frictional torque, slippage between sprocket 2 and the fingertip unit does not occur; thus, the fingertip unit rotates together with sprocket 2. The condition for the fingertip unit to rotate without slippage between sprocket 2 and the fingertip unit is expressed as:

$$\tau_{rc2} \leq \tau_{pre}^{st.max} + \tau_{g1} \quad (15)$$

where $\tau_{pre}^{st.max}$ is the maximum static frictional torque. In contrast, the condition to generate the relative rotation with slippage between sprocket 2 and the fingertip unit is given by

$$\tau_{rc2} > \tau_{pre}^{st.max} + \tau_{g1} \quad (16)$$

*3) Conditions for the desired mode-switching order*

This section describes the conditions for realizing the desired mode-switching order: finger-bending, grasping, and pull-in modes. In the initial state, where the finger is extended, $f_{tip}= 0$ because there is no contact between the fingertip and the object. $\tau_{stopper}= 0$ because there is no contact between the stopper in the fingertip unit and the finger body; thus, from (14), $\tau_{rc2}$ is also zero. In this case, condition (8) for activating the finger-bending mode in the initial state is expressed by

$$|\alpha_{st}r_{dr}\Delta f| > |\Delta\tau| \quad (17)$$

The lead angle of the thread $\lambda$ that determines $\alpha_{st}$ is the design parameter. By tuning $\lambda$ to satisfy (17), the finger-bending mode can be activated first. When the fingertip unit rotates to the point where it is perpendicular to the finger body and the stopper in the fingertip unit contacts the finger body, $\tau_{stopper}$ increases. Accordingly, $\tau_{rc2}$ increases. Because $f_{tip} = 0$ in this state, condition (11) for the translation of sprocket 1 is rewritten as

$$|(\tau_{rc2} + \Delta\tau)/r_{dr}\alpha_{st}| > |\Delta f| \quad (18)$$

If (15) and (18) are satisfied, sprocket 1 is translated without relative rotation between sprocket 2 and the fingertip unit. Thus, the grasping mode is activated after the finger-bending mode. This condition can be satisfied by tuning $\tau_{pre}^{st.max}$, which is determined by the magnitude of the tightening torque, $T_{scw}$, of the screw in SMS mechanism B. Thus, $T_{scw}$ is the tuning parameter. After switching to the grasping mode, the following force and moment balances in sprocket 1 are satisfied:

$$|f_{SM}|_{f_{\text{fri}}=\mu_{kn}f_n} = |f_{tip} + \Delta f| \quad (19)$$
$$|\tau_{SM}|_{f_{\text{fri}}=\mu_{kn}f_n} = |\tau_{rc2} + \Delta\tau| \quad (20)$$

where $\mu_{kn}$ is the kinetic friction coefficient. From (6), the relationship between the balances is rewritten as

$$|f_{tip} + \Delta f| = \left|\frac{1}{r_{dr}\alpha_{kn}}(\tau_{rc2} + \Delta\tau)\right| = \left|\frac{1}{r_{dr}\alpha_{kn}}\left(\frac{\tau_m}{2} + \Delta\tau\right)\right| \quad (21)$$
$$\alpha_{kn} = (\mu_{kn}\cos\lambda + \sin\lambda)/(\cos\lambda - \mu_{kn}\sin\lambda) \quad (22)$$

From (21), $\tau_{rc2}$ increases as $f_{tip}$ increases. If (16) is satisfied by an increase in $\tau_{rc2}$ due to further motor rotation and an increase in $f_{tip}$, the pull-in mode is activated. During the pull-in mode, $\tau_{rc2}$ is given by

$$\tau_{rc2} = \tau_{pre}^{kn} + \tau_{g1} \quad (23)$$

where $\tau_{pre}^{kn}$ is the kinetic frictional torque. In this manner, the desired motion-switching order can be realized. As shown in Fig. 1(a), it is desirable to further switch the mode to the finger-bending mode, resulting in finger extension. If the motor is rotated in the reverse direction while grasping the object, the contact between the stopper in the fingertip unit and the finger body is lost and $\tau_{rc2}$ ($= \tau_{stopper}$) becomes zero. From (8), the condition for activating the finger-bending mode in this case is:

$$|\alpha_{st}r_{dr}(f_{tip} + \Delta f)| > |\Delta\tau| \quad (24)$$

This condition can be satisfied if the design parameters are set such that (17) can be satisfied. The finger-extending motion in the finger-bending mode continues until the stopper contacts the other side of the finger body. After the contact, $\tau_{rc2}$ increases owing to the increase in $\tau_{stopper}$, and (11) is satisfied. Then, the finger-opening motion in the grasping mode is activated, and the object can be released. Hence, the motion shown in Fig. 1(a) can be realized.

*B. Validation*

This section presents the experimental validation of the analysis described in Section III.A. First, SMS mechanism A was evaluated using the experimental setup shown in Fig. 7. This experiment evaluated the behavior of the finger during the finger-bending and grasping modes. The fingertip unit was attached to sprocket 2 such that relative rotation between them did not occur and the pull-in mode was not activated. The torque applied from roller chain 2 to sprocket 1 ($\tau_{rc2}$), tip force ($f_{tip}$), rotational angle of sprocket 1 ($\theta_{sp1}$), and translational distance of the finger ($d_f$) were measured. $d_f$ corresponds to the translational distance of sprocket 1. Force and torque sensors (IMADA, ZTS-500N, HTGS-10N) were installed between sprocket 2 and the fingertip unit to measure $\tau_{rc2}$ and $f_{tip}$. The torque measured by the sensor corresponds to $\tau_{rc2}$, while the force measured corresponds to $f_{tip}$. Camera 1 (Logicool, C1000eR) was placed behind the finger to capture marker 1 mounted on sprocket 2, and the rotational angle of sprocket 2 ($\theta_{sp2}$) was measured using image processing. Let $\theta_f$ be the rotational angle of the fingertip unit. Note that $\theta_{sp1} = \theta_{sp2} = \theta_f$ because sprockets 1 and 2 were connected via roller chain 2, and the fingertip unit and sprocket 2 were connected in such a way that relative rotation was not possible, as described above. Camera 2 (Logicool, C1000eR) was placed on the upper side of the finger to capture marker 2 mounted on the upper surface of the finger and derive $d_f$ using image processing. The target object was firmly fixed to the environment such that it would not move upon contact with the finger. The finger body was initially set such that the distance between the fingertip and object was 10 mm. The motor torque $\tau_m$, calculated from the motor current, was also measured. Note that the value of $\tau_m$ in this experiment was calculated as twice the actual measured value because the experiment was conducted with only one finger. The motor was rotated at a constant velocity of 14 deg/s until the calculated $\tau_m$ reached 0.8 Nm. As described in Section III.A.3, the lead angle $\lambda$ of the threads of the drive shaft and sprocket 1 is the key design parameter for determining the behavior of SMS mechanism A. The experiment was conducted under three conditions with $\lambda$ values of 20°, 30°, and 60°. The experiments were conducted 10 times for each $\lambda$ condition. The results are shown in Fig. 8. The desired behavior was achieved when $\lambda = 30°$ (Fig. 8(a)). $\theta_{sp1}$ initially increased and the fingertip rotated without translation of the finger (i.e., $d_f$ remained at 0 mm). This result indicates that (17) was satisfied in this case. When $\theta_f$ reached 90°, where the stopper in the fingertip unit contacted the finger body, $\tau_{rc2}$ begun to increase owing to the increase in $\tau_{stopper}$. (18) was satisfied by the increase in $\tau_{rc2}$, and the finger was translated (i.e., $d_f$ increased). In this manner, the self-motion switching from the finger-bending mode to the grasping mode was achieved. When the translation of the finger stopped due to the contact between the fingertip and the object, $\tau_{rc2}$ begun to increase further as $f_{tip}$ increased. A further increase in $\tau_{rc2}$ is used to activate the pull-in mode in SMS mechanism B. When $\lambda = 20°$ (small), unstable behavior was observed in which the rotation and translation of sprocket 1 were intermixed (Fig. 8(b)). This is because $\alpha_{st}$ was small due to the small $\lambda$, and the cases that satisfy (17) and those that do not satisfy (17) occurred in a mixture. When $\lambda = 60°$ (large), the desired behavior was achieved (Fig. 8(c)); however, the obtained $f_{tip}$ was small because $\alpha_{kn}$ was large due to the large $\lambda$ in (21). In all cases, the behavior of $\tau_m$ was the same as that of $\tau_{rc2}$. Meanwhile, the time needed for $d_f$ to reach 10 mm increases as $\lambda$ decreases, indicating a tradeoff relationship between the finger motion speed and the tip force. These results validated the analysis of switching mechanism A.

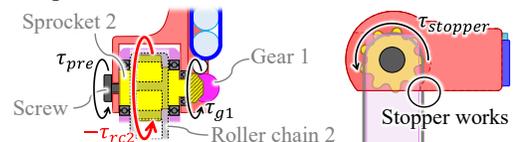

**Fig. 6.** Mechanical relationship in SMS mechanism B

Next, SMS mechanism B was evaluated using the experimental setup shown in Fig. 9. To evaluate the mode switching between the finger-bending and pull-in modes in SMS mechanism B, the finger without the grasping function was used. Sprocket 1 was driven directly by the motor, and the finger body was fixed to the environment. The rotational angle of the fingertip ($\theta_f$), rotational angle of sprocket 2 ($\theta_{sp2}$), and torque applied to sprocket 2 from roller chain 2 ($\tau_{rc2}$) were measured. The torque sensor was installed between the motor and sprocket 1. The torque measured by the torque sensor (IMADA, HTGS-10N) corresponds to $\tau_{rc2}$. To measure $\theta_f$ and $\theta_{sp2}$, the camera (Logicool, C1000eR) was placed in front of the finger. Markers 1 and 2 were attached to the fingertip frame and sprocket 2, respectively, as shown in Fig. 9(b). Note that the belt-conveyor mechanism was removed to install marker 2. By using image processing, $\theta_f$ and $\theta_{sp2}$ were obtained. As described in Section III.A.3, the tightening torque of the screw ($T_{scw}$), which determines $\tau_{pre}^{st.max}$, is the key tuning parameter for achieving the desired mode-switching order. The

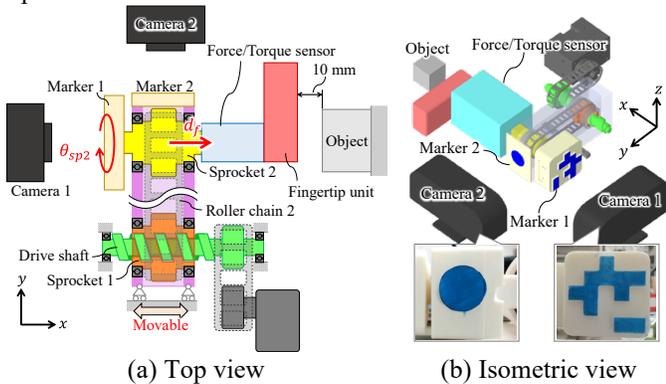

(a) Top view    (b) Isometric view
**Fig. 7.** Setup for the evaluation of SMS mechanism A

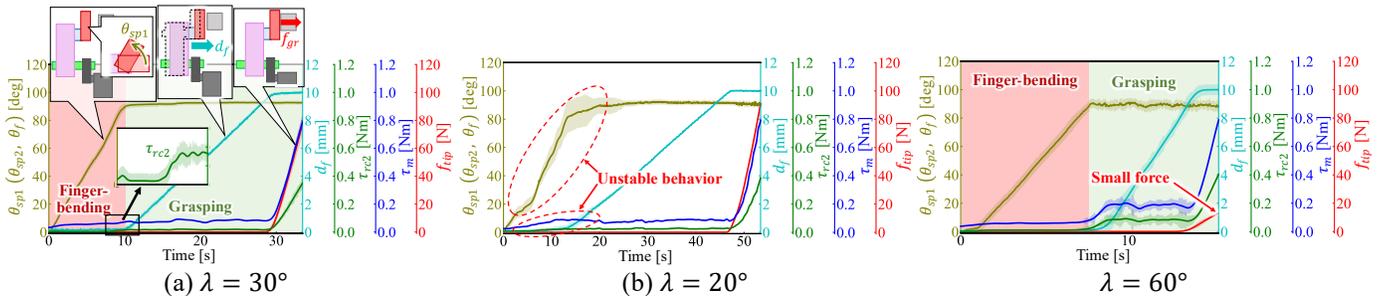

(a) $\lambda = 30°$    (b) $\lambda = 20°$    $\lambda = 60°$
**Fig. 8.** Result of the evaluation of self-motion switching mechanism A experiments were conducted under three conditions, where $T_{scw}$ was set to 0.0, 0.46, and 0.75 Nm. The experiments were conducted 10 times for each condition of $T_{scw}$. The motor rotated 180° at a constant velocity of 14 deg/s. Fig. 10 shows the results of $\theta_f$, $\theta_{sp2}$, and $\tau_{rc2}$. When $T_{scw}$ was large (0.46 and 0.75 Nm), the fingertip unit and sprocket 2 initially rotated together (Figs. 10 (a) and (b)). After the fingertip unit and sprocket 2 rotated 90°, where the stopper in the fingertip unit worked, $\theta_f$ remained at 90° while $\theta_{sp2}$ continued to increase. Thus, relative rotation between the fingertip unit and sprocket 2 was generated. The relative rotation activates the pull-in function if the belt-conveyor mechanism is installed. Hence, the desired mode-switching order was achieved. When the switching was activated, $\tau_{rc2}$ reached its peak value because the frictional state became the maximum static frictional state. The peak value when $T_{scw}$ = 0.75 Nm was larger than that when $T_{scw}$ = 0.46 Nm, validating (16) because $\tau_{pre}^{st.max}$ increases as $T_{scw}$ increases. During the pull-in mode, $\tau_{rc2}$ was constant, as expressed in (23). When $T_{scw}$ was small (0.0 Nm), the fingertip unit was not rotated (i.e., $\theta_f$ remained at 0°). The relative rotation between sprocket 2 and the fingertip unit occurred in the initial state. Under this condition, the threshold ($\tau_{pre}^{st.max} + \tau_{g1}$) for generating relative rotation was extremely small. Thus, (16) was initially satisfied, and the pull-in mode was activated before the finger-bending mode. These results validate the analysis described in Section III.A. $T_{scw}$ should exceed a certain value to avoid the behavior shown in Fig. 10(c). Setting a large $T_{scw}$ results in a large $\tau_{pre}^{st.max}$, which leads to a large generable tip force in the grasping mode from (15) and (21). However, $T_{scw}$ should be within the range where the motor torque $\tau_m|_{\tau_{rc2}=|\tau_{scw}^{st.max}+\tau_{g1}|}$ is smaller than its maximum, to activate the pull-in mode.

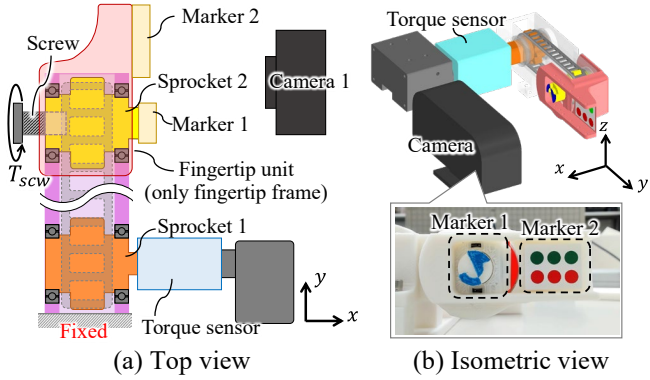

Fig. 9. Setup for the evaluation of SMS mechanism B

In summary, SMS mechanism A activates the finger-bending mode from the initial state and switches to the grasping mode. SMS mechanism B prevents the activation of the pull-in mode before the activation of the finger-bending mode. As shown in Fig. 8, $\tau_{rc2}$ increased further with an increase in $f_{tip}$ after the translation of the finger stopped. As can be estimated from the values of $\tau_{rc2}$ shown in Figs. 8 and 10, a further increase in $\tau_{rc2}$ triggers the activation of the pull-in mode. The results demonstrate that by tuning $\lambda$ and $T_{scw}$, it is possible to achieve the desired motion behavior of switching in the order of finger-bending, grasping, and pull-in modes.

## IV. CONTROL METHODOLOGY

This section presents the control methodology of the developed gripper. The manipulated variables are the motor torque $\tau_m$ and rotational angle of the motor $\theta_m$. The control targets are the finger posture $\theta_f$ in the finger-bending mode, grasping (tip) force $f_{tip}$ in the grasping mode, and translational distance $d_{PI}$ of the pulled-in object in the pull-in mode. The control of the gripper also requires the detection of the points at which the mode switches: the point of the switch from the finger-bending mode to the grasping mode, and the point from the grasping mode to the pull-in mode. Fig. 11 shows $\tau_m$ and the state of the gripper mode while the motor rotated at a constant velocity of 14 deg/s and the target object (width: 35 mm) was grasped. $\lambda$ was set to 30° and $T_{pre}$ was set to 0.46 Nm to achieve the desired mode-switching order and to maximize the generable tip force while avoiding vibrated motion behavior. The finger was initially bent according to the motor rotation ($\theta_m$). After $\theta_m$ reached 90°, the grasping mode was activated and $\tau_m$ increased as the grasping force increased. When $\tau_m$ reached the peak value, the pull-in mode was activated. Based on the result, the methodology to detect the switching points and to control $\theta_f$, $f_{tip}$, and $d_{PI}$ is proposed, as shown in Fig. 12. In this methodology, the motor rotates at a constant velocity while monitoring $\tau_m$ and $\theta_m$. The target finger posture $\theta_f^{target}$ can be achieved by controlling the motor angle as $\theta_m(=\theta_f) \to \theta_f^{target}$ $(0 \leq \theta_f^{target} \leq 90°)$. The switching point from the finger-bending mode to the grasping mode can be detected as the point where $\theta_m(=\theta_f)$ becomes 90°. In the grasping mode, $f_{tip}$ is determined by $\tau_m$ from (21). If $\Delta f$ and $\Delta \tau$ are negligibly small, the $\tau_m$ ($\tau_m^{target}$) that provides the target $f_{tip}$ ($f_{tip}^{target}$) is:

$$|\tau_m^{target}| = |2r_{dr}\alpha_{kn}f_{tip}^{target}| \quad (25)$$

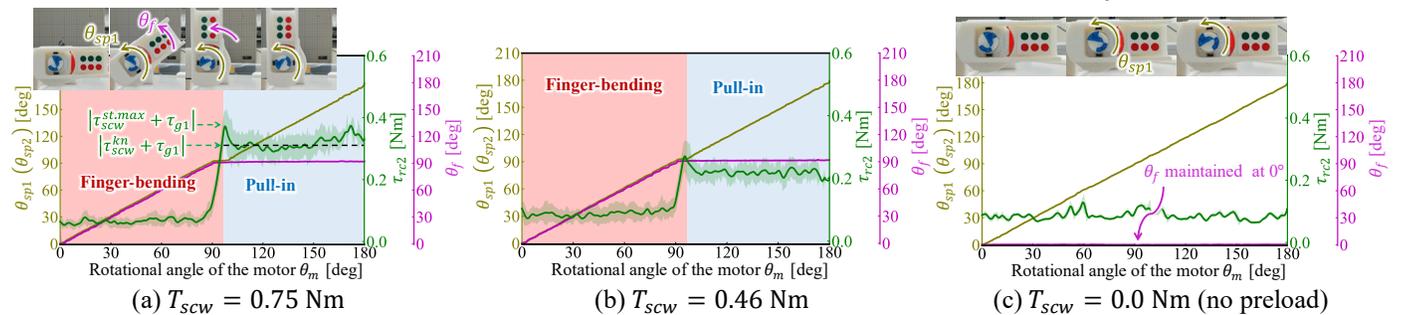

(a) $T_{scw} = 0.75$ Nm  (b) $T_{scw} = 0.46$ Nm  (c) $T_{scw} = 0.0$ Nm (no preload)

Fig. 10. Result of the evaluation for self-motion switching mechanism B

The tip force can be controlled to be $f_{tip}^{target}$ by controlling $\tau_m \rightarrow \tau_m^{target}$. To detect the point of switching to the pull-in mode, the threshold ($\tau_m^{th}$) of $\tau_m$ is defined. As shown in Fig. 11, the switching to the pull-in mode is activated when $\tau_m$ reaches its peak value. This activation can be detected by setting $\tau_m^{th}$ to be slightly smaller than the peak value for stable detection. Hence, $f_{tip}^{target}$ can be set within the range where $\tau_m^{target} < \tau_m^{th}$. The target $d_{PI}$ ($d_{PI}^{target}$) can be realized by controlling the motor angle after the switching to the pull-in mode is detected. Let $\theta_m^{g \rightarrow p}$ be the motor angle when the switching to the pull-in mode is activated and $\theta_{m_{PI}} = \theta_m - \theta_m^{g \rightarrow p}$. Assuming that no slippage occurs between the belt of the belt-conveyor mechanism in the fingertip unit and the grasped object, $\theta_{m_{PI}}$ ($\theta_{m_{PI}}^{target}$), which provides $d_{PI}^{target}$, is given by

$$\theta_{m_{PI}}^{target} = d_{PI}^{target} / (\varepsilon r_{\text{Rolloer}}) \quad (26)$$

where $r_{\text{Rolloer}}$ is the radius of the roller in the belt conveyor mechanism, and $\varepsilon$ is the gear ratio obtained from the number of teeth in the bevel gear part of sprocket 1, gears 1 and 2, and the gear of the roller. The pull-in amount of the object can be controlled to be $d_{PI}^{target}$ by controlling $\theta_{m_{PI}} \rightarrow \theta_{m_{PI}}^{target}$. The proposed methodology is validated in the next section.

## V. Grasping test

### A. Grasping in confined spaces

This section describes the grasping tests for picking up objects from confined spaces. The target situations in which conventional parallel-jaw grippers cannot grasp an object are shown in Fig. 13. In cases 1 and 2, the target object was placed in an open-top box in the shelf, and the operating range of the gripper body was limited (Figs. 13 (a) and (b)). In case 1, obstacle boxes were also placed inside the box, limiting the movable range of the target object. The result of the grasping test in this case is shown in Fig. 1(a). Using the control methodology, the object was translated 50 mm upward in the pull-in mode. As described in Section I, the picking was successful. In case 2, the object could not be picked up if it was rotated, as in case 1. The gripper picked up the object by not extending its fingers after the pull-in operation, as shown in Fig. 14. The object was pulled in 40 mm using the control methodology. In cases 3 and 4, the object was placed in a box with a partially opened top surface; note that it was not placed directly under the opening (Fig. 13 (c) and (d)). In case 3, the fingers were bent 90° using the finger-bending mode to grasp the object placed at the bottom of the box. Subsequently, the object was grasped and pulled in 40 mm. After that, the fingers were extended to avoid collision with the box. The picking was successful (Fig. 15). In case 4, the object posture differed from that in case 3. Re-grasping was adopted to grasp the object such that the longitudinal direction of the fingertip unit was aligned with the longitudinal direction of the object, i.e., the edge with the length of 35 mm (Fig. 16). The gripper rotated the grasped object by extending its fingers, as in case 3. The object was then placed in the box by opening its fingers through the reversed motor rotation. The grasping mode was activated because the finger was extended. The object posture then became the same as the initial posture in case 3. The remainder of the picking procedure was the same as in case 3. Thus, the picking process was successful. These results demonstrate the effectiveness of the developed gripper for grasping in confined spaces.

### B. Grasping objects on a table

This section describes grasping tests that involve grasping various objects placed on a table. The target objects included a small die (side length of 15 mm), heavy bottle (1.0 kg), fragile egg, spherical golf ball, and deformable toy doll. The target motor torques $\tau_m^{target}$, which gives the desired tip force of 28 N for objects other than the egg and 9 N for the egg, were derived

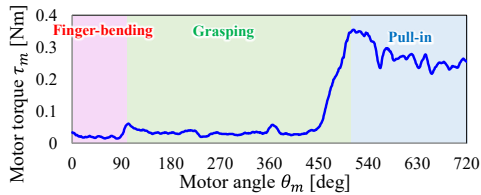

**Fig. 11.** Behavior of the gripper and motor torque $\tau_m$

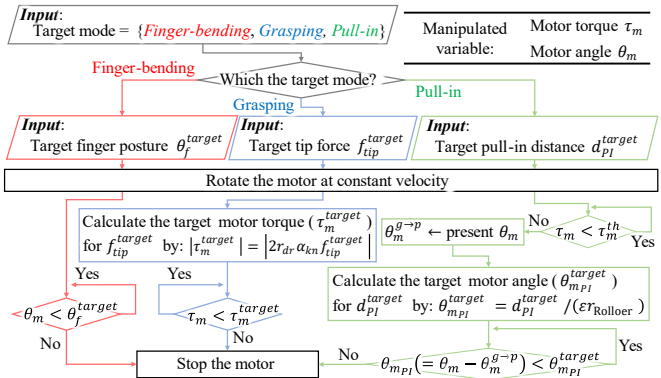

**Fig. 12.** Control methodology

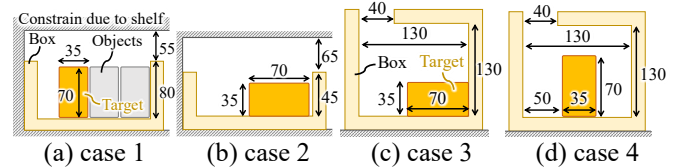

(a) case 1 (b) case 2 (c) case 3 (d) case 4
**Fig. 13.** Target situations (unit: mm)

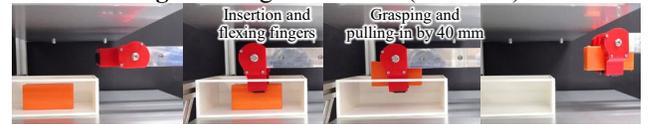

**Fig. 14.** Snapshots of the grasping test in case 2

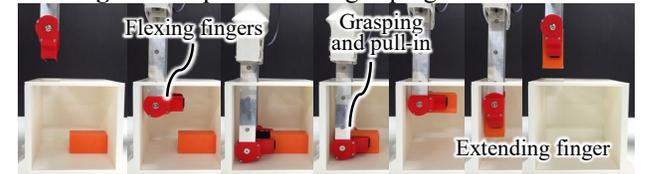

**Fig. 15.** Snapshots of the grasping test in case 3

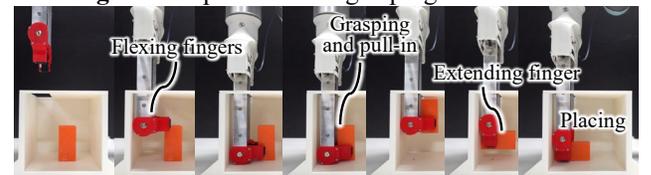

**Fig. 16.** Re-grasping process (first half) in case 4

from (25). $r_{dr}$ was 5 mm and $\mu_{kn}$ was assumed to be 0.3. The calculated $\tau_m^{target}$ values were 0.3 Nm for objects other than the egg and 0.1 Nm for the egg. All objects were successfully grasped (see the video clip). In addition, the gripper released the object by only rotating the motor in the reverse direction. As shown in Fig. 17, the gripper grasped the object with bending its fingers. When rotating the motor in the reverse direction in this state, the grasping mode is activated first, not the finger-bending mode. This is because frictional torque is generated on the contact area between the fingertip and the object, preventing the rotation of the fingertip, i.e., the activation of the finger-bending mode. Thus, the finger-opening motion in the grasping mode is activated. If the fingers open until the frictional torque is removed, the finger-bending mode is activated, and the fingers are extended. After finger extension, the grasping mode is activated, and the fingers are opened.

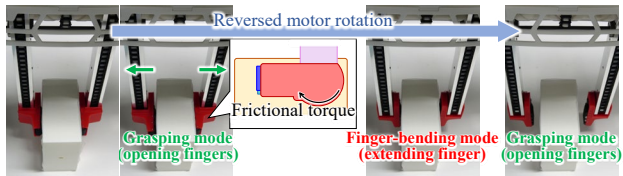

**Fig. 17.** Releasing motion using the reversed motor rotation

## VI. CONCLUSION

This study developed a novel robotic gripper with grasping, finger-bending, and pull-in modes for grasping in confined spaces. The gripper is equipped with two SMS mechanisms to achieve these three modes using a single motor. SMS mechanism A switches between the grasping mode and the other two modes. SMS mechanism B switches between the finger-bending and pull-in modes. The mode switching occurs automatically when the motion being generated is prevented by a disturbance, including contact with the stopper or an increase in tip force. The SMS mechanisms reduce the number of motors, resulting in a compact gripper design with thin fingers (the width of the finger was 40 mm in the initial state). The installation of SMS mechanism B enlarged the finger compared to the simple thin fingers used in, e.g., [4]–[11], which is disadvantageous for inserting the fingers into extremely narrow spaces. Finger downsizing could be achieved by using small-scaled parts. However, the small scaling would cause a decrease in the stiffness and durability of the finger; thus, the finger should be designed considering this trade-off relationship. The three functional modes and thin fingers enable the gripper to pick up an object from a confined space. The gripper mechanism was analyzed, and the design and tuning parameters were introduced to achieve the desired mode-switching order. A control methodology was also proposed to control the bending angle of the fingertips, tip forces, and pull-in amount while activating the desired modes. The validity of the methodology was demonstrated through grasping tests. The performance of the gripper was experimentally evaluated. The results demonstrated that the payloads in the grasping and pull-in modes were 30.8 N and 29.4 N, respectively (see the video clip for the performance details). Although the modes are normally activated in the determined order, the order can be changed in the proposed mechanism by adjusting the parameters or adding intentional disturbances. This mechanism enables the gripper to release the grasped object by only rotating the motor in the reverse direction. In the grasping tests, the robotic manipulator equipped with the gripper was operated using manual teaching. To realize a fully automated picking-up system, additional sensors would be needed. Our future work will involve the evaluation of the effectiveness of the gripper in other situations, the development of a method to optimize the design parameters, and the integration of sensors for full automation.